\documentclass{ncjms}

\usepackage{verbatim}
\usepackage{subcaption}
\usepackage{listings}
\usepackage{times}
\usepackage{epsfig}
\usepackage{amssymb}
\usepackage{tabularx} 
\usepackage{amsmath}  
\usepackage{graphicx} 
\usepackage{adjustbox}
\usepackage{caption}
\usepackage{enumitem}
\usepackage[symbol]{footmisc}

\makeatletter
\newcommand\footnoteref[1]{\protected@xdef\@thefnmark{\ref{#1}}\@footnotemark}
\makeatother

\usepackage{geometry}
\usepackage{cite} 

\usepackage{tikz}
\usetikzlibrary{shapes,arrows, fit}

\tikzstyle{decision} = [diamond, draw, fill=uncwblue!50, 
text width=6em, text badly centered, node distance=3cm, inner sep=0pt]
\tikzstyle{block} = [rectangle, draw, fill=uncwblue!50, 
text width=6em, text centered, rounded corners, minimum height=3em]
\tikzstyle{block2} = [rectangle, draw, fill=uncwyellow!20, 
text width=2em, text centered, rounded corners, minimum height=2em]
\tikzstyle{block3} = [rectangle, draw, fill=uncwyellow!20, 
text width=6em, text centered, rounded corners, minimum height=2em]
\tikzstyle{line} = [draw, -latex']
\tikzstyle{cloud} = [draw, ellipse,fill=uncwteal!50, node distance=3cm,
minimum height=4em]
\tikzstyle{black} = [draw, regular polygon,regular polygon sides=10, node distance=3cm, minimum height=5em]
\tikzstyle{white} = [draw, circle,node distance=3cm, minimum height=5em]

\definecolor{uncwblue}{RGB}{0, 51, 102}
\definecolor{uncwteal}{RGB}{0, 112, 115}
\definecolor{uncwyellow}{RGB}{255, 215, 0}

\begin{document}
	\title{Image Pre-processing Using OpenCV Library on MORPH-II Face Database}
	\titlerunning{Pre-processing Using OpenCV}   
	
	\author{B. Yip, R. Towner, T. Kling, C. Chen, and Y. Wang}
	\address[C. Chen]{Department of Mathematics and Statistics, The University of North Carolina Wilmington, Wilmington, NC 28403, USA}
	\email[Corresponding author]{chenc@uncw.edu}
	\authorsrunning{B.~Yip et al.}  

	\begin{abstract}
		This paper outlines the steps taken toward pre-processing the 55,134 images of the MORPH-II non-commercial dataset. Following the introduction, section two begins with an overview of each step in the pre-processing pipeline. Section three expands upon each stage of the process and includes details on all calculations made, by providing the OpenCV functionality paired with each step. The last portion of this paper discusses the potential improvements to this pre-processing pipeline that became apparent in retrospect. 
\end{abstract}
		
		
		
		
		\date{\today}
		
		\maketitle
		
		%
		%
			
\section{Introduction}
			The MORPH data is one of the largest publicly available longitudinal face databases \citep{ricanek2006morph}. Since its first release in 2006, it has been cited by over 500 publications. Multiple versions of MORPH have been released, but for our face image analysis study, we use the 2008 MORPH-II non-commercial release. The MORPH-II dataset includes 55,134 mugshots with longitudinal spans taken between 2003 and late 2007. For each image, the following metadata is included: subject ID number, picture number, date of birth, date of arrest, race, gender, age, time since last arrest, and image filename. Because of its size, longitudinal span, and inclusion of relevant metadata, the MORPH-II dataset is widely utilized in the field of computer vision and pattern recognition, including a variety of race, gender, and age face imaging tasks.

\begin{figure}[htbp]
				\begin{subfigure}{0.35\textwidth}
					\includegraphics[width=\textwidth]{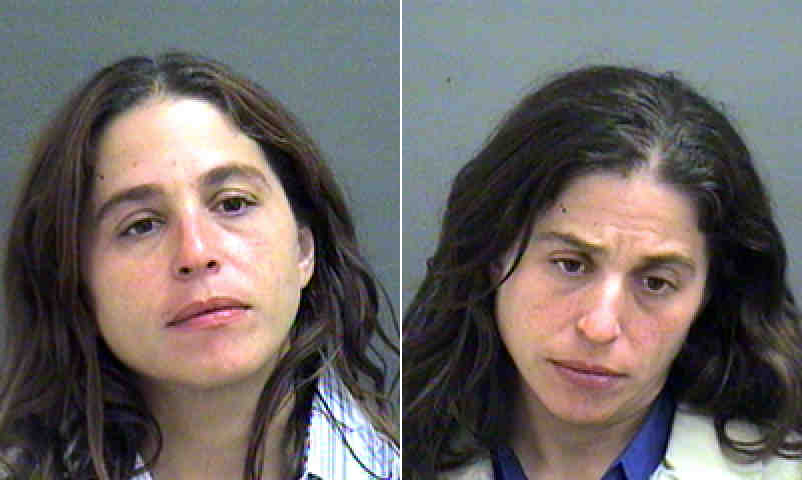}
                    \label{variety_headtilt}
                    \caption{Variation in head tilt.}
				\end{subfigure}%
                \hspace{0.25cm}
				\begin{subfigure}{0.35\textwidth}
					\includegraphics[width=\textwidth]{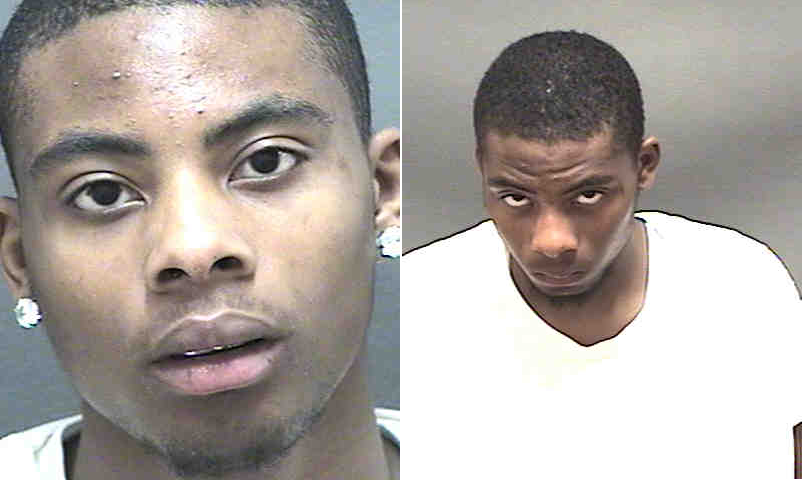}
                    \label{variety_distance}
                    \caption{Variation in camera distance.}
				\end{subfigure}
                
                \vspace{0.25cm}
                
				\begin{subfigure}{0.35\textwidth}
					\includegraphics[width=\textwidth]{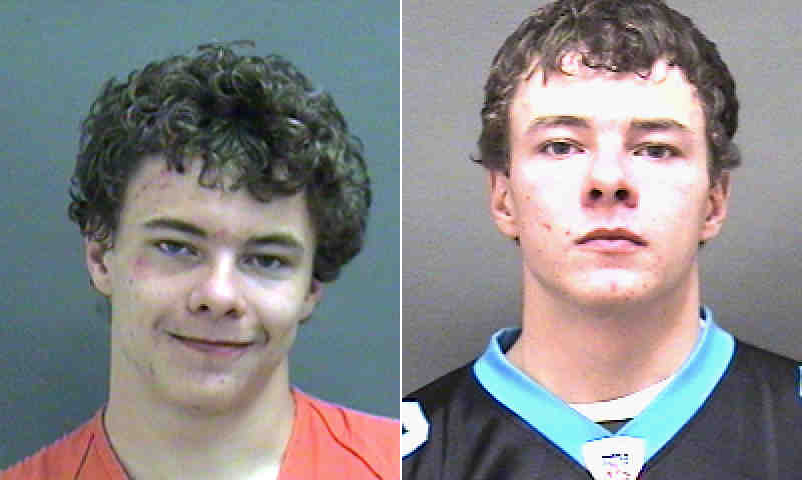}
                    \label{variety_illumination}
                    \caption{Variation in illumination.}
				\end{subfigure}%
                \hspace{0.25cm}
				\begin{subfigure}{0.35\textwidth}
					\includegraphics[width=\textwidth]{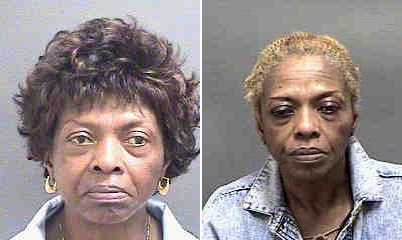}
                    \label{variety_female}
                    \caption{Variation in overall appearance.}
				\end{subfigure}%
				\caption{Different types of variation present in the MORPH-II dataset.}
\label{variety}
\end{figure}

			However, despite the fairly standard format of police photography, many of the images vary greatly in terms of head-tilt, camera distance, and illumination. A great number of the images also contain large, empty backgrounds or excess occlusion that add a corresponding amount of noise to the data. This longitudinal dataset had an average of approximate 4 images per subject and some appearances varied greatly from one image to the next. Preliminary results showed women having an increased overall variation in their images due to changes in makeup and hairstyle. Figure \ref{variety} showcases some variety found during initial examination of the image dataset.
			
			Consequently, the pre-processing step is crucial for the image analysis on the MORPH-II dataset. For our purposes, we utilized the Open Source Computer Vision 2 (OpenCV) library in Python to extract the face from each mugshot using the image vectors \citep{opencv_library}. The stages of this process are outlined in section 2.

		\section{Procedural Overview}
\begin{figure}[hb]
\centering
    \begin{minipage}[t]{.35\linewidth}
      \centering
      \includegraphics[width=.6\textwidth]{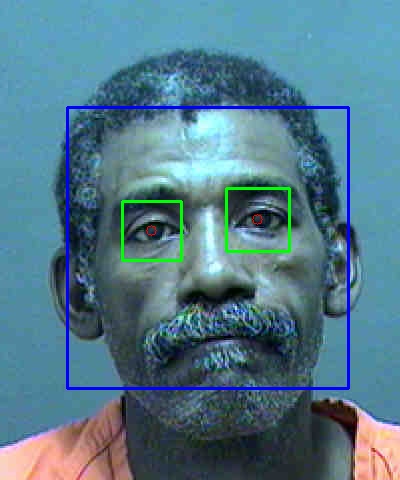}\\
      \textbf{Initial face and eyes detection.}
    \end{minipage}%
    \hspace{-1cm}
    \begin{minipage}[t]{.35\linewidth}
      \centering
      \includegraphics[width=.6\textwidth]{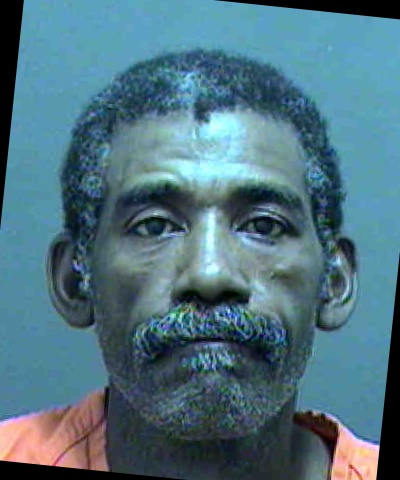}\\
      \textbf{Rotation.}
    \end{minipage}%
    \hspace{-1cm}
    \begin{minipage}[t]{.35\linewidth}
      \centering
      \includegraphics[width=.6\textwidth]{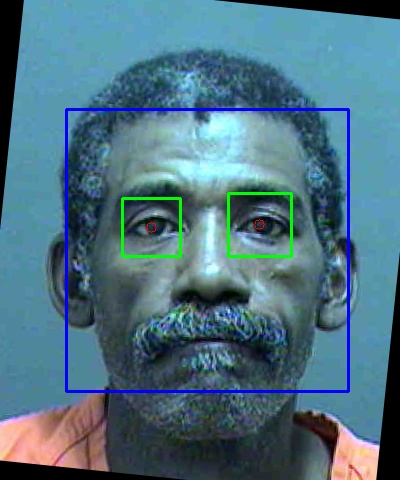}\\
      \textbf{Face and eye re-detection.}
    \end{minipage}
    
    \vspace{.2cm}
    \centering
    \begin{minipage}[t]{.35\linewidth}
      \centering
      \includegraphics[width=.6\textwidth]{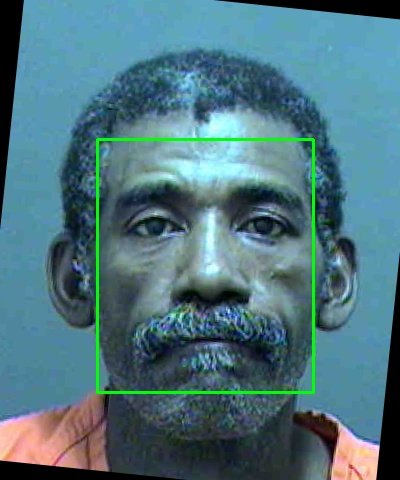}\\
      \textbf{Cropping and scaling}
    \end{minipage}%
    \hspace{-1cm}
    \begin{minipage}[t]{.35\linewidth}
      \centering
      \includegraphics[width=.6\textwidth]{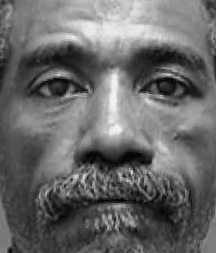}\\
      \textbf{Pre-processed image.}
    \end{minipage}
    \caption{Stages of the pre-processing pipeline with successful face and eye detection}
        \label{pipeline1}
  \end{figure}

\begin{figure}[t!]
			\label{ImageProcessPipeline}
				\begin{subfigure}{0.25\textwidth}
					\includegraphics[width=\textwidth]{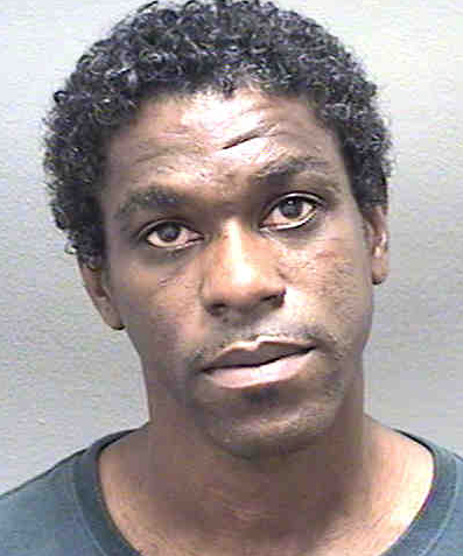}
                    \label{}
                    \caption{Original image.}
				\end{subfigure}%
                \hspace{0.25cm}
				\begin{subfigure}{0.25\textwidth}
					\includegraphics[width=\textwidth]{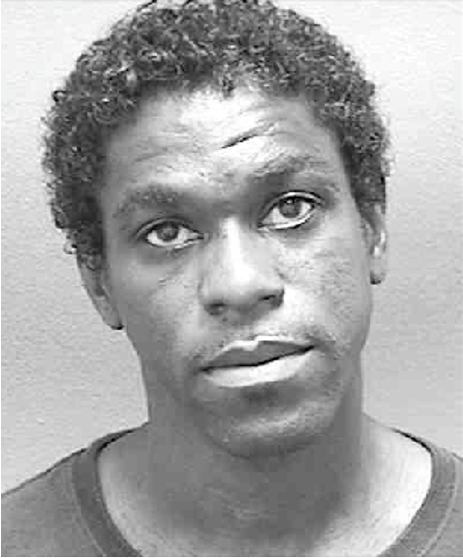}
                    \label{}
                    \caption{Grayscale image.}
				\end{subfigure}
                \hspace{0.25cm}
				\begin{subfigure}{0.25\textwidth}
					\includegraphics[width=\textwidth]{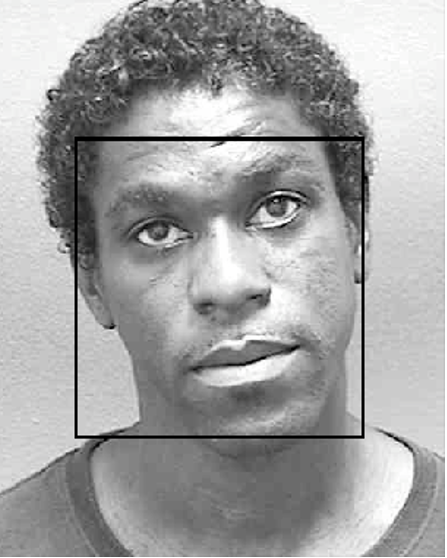}
                    \label{}
                    \caption{Initial face detection.}
				\end{subfigure}
                
                \vspace{0.25cm}
                
				\begin{subfigure}{0.25\textwidth}
					\includegraphics[width=\textwidth]{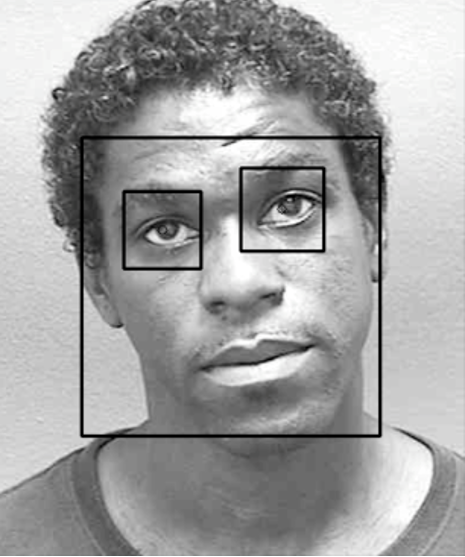}
                    \label{}
                    \caption{Eye detection.}
				\end{subfigure}%
                \hspace{0.25cm}
				\begin{subfigure}{0.25\textwidth}
					\includegraphics[width=\textwidth]{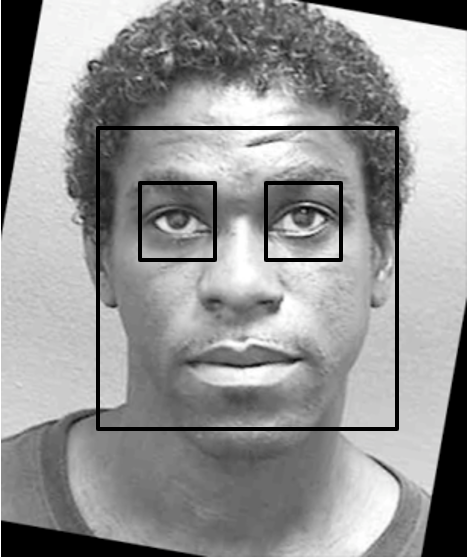}
                    \label{}
                    \caption{Rotation \& re-detection.}
				\end{subfigure}
                \hspace{0.25cm}
				\begin{subfigure}{0.25\textwidth}
					\includegraphics[width=\textwidth]{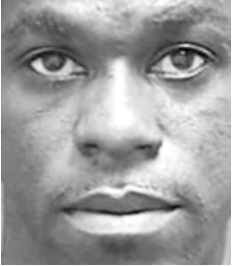}
                    \label{}
                    \caption{Cropping and scaling.}
				\end{subfigure}
               \caption{Face preprocessing pipeline with successful face and eye detection.}
               \label{pipeline2}
			\end{figure}

			This section provides a global description for the six stages of our pre-processing algorithm. The premise is to minimize image noise by the use of bounding boxes around necessary region of interest (ROI). Both Figures \ref{pipeline1} and \ref{pipeline2} are visual representations of each step and what is accomplished, from different prospects.
            
\noindent \textbf{Note:} While the intermediate stages of the process in Figure \ref{pipeline1} are shown with color images, all computer vision tasks were done only on the grayscale versions of each image (converted with OpenCV).\\

			\subsection{Grayscale}
				Research in computer vision showed converting images to grayscale increased the accuracy of locating the necessary facial features as it reduced the effect illumination variance had on the images. Doing so was the first introduction to the OpenCV library. For each image, we utilized the OpenCV function $cv2.cvtColor(src, channel)$ where $src$ is the input image and $channel$ represents the color channel for the output image. In our case we used $Color_BGR2Gray$. This results in the new image pixel value, $Y$, where $Y = 0.299 \cdot R + 0.587 \cdot G + 0.114 \cdot B$, and $R$, $G$, $B$ represent Red, Green and Blue respectively. The values of $R$, $G$, $B$, are from the original image pixel. Effects are shown in Figure \ref{pipeline2}(B).

			\subsection{Face Detection}
				The initial face detection step located and marked the position of a face within an image, which can be seen in Figure \ref{pipeline2}(C). This eliminates backgrounds and hairstyles, which are image properties that are not useful for computer vision tasks. This procedure increased the accuracy of future detection steps. If a face was not successfully detected, the image was stored in a face not found (fnf) folder for manual detection later on. Both face and eye detection steps were accomplished using Haar-Feature based cascade classifiers from OpenCV. The function used was $cv2.cascadeClassifier.detectMultiScale(src, sf, mn)$ where $src$ is the input image, $sf$ is the scale factor at each image scale, and $mn$ is the minimum amount of neighbors each candidate face rectangle should acquire.
                
               Note that: Both face and eye detection were done using the the Haar-feature based Cascade Classifiers from OpenCV (the .xml files can be obtained from the OpenCV GitHub repository). For our purposes, we only had to adjust the parameters of the OpenCV detection function to locate the face in each image. Eye detection, however, required additional steps.

			\subsection{Eye Detection}
            	We implemented a very similar algorithm for locating the eyes, illustrated in Figure \ref{pipeline2}(D). After the face is detected, the eyes are located within the region of interest (ROI) determined by the face (i.e. within the bounding box for the face), and the eye centers are computed as the center of the bounding box for each eye. The new domain and range of the image matrix come from the bounding box around a face that was found successfully. We then marked bounding boxes around each eye with the same Haar cascade function from OpenCV in section 2.2 with different parameter values to account for the smaller scope. 
                
                In many cases, wrinkles, shadows, and other facial blemishes were detected as eyes. To account for this, we implemented two conditions to eliminate as many of the incorrect eye detections as possible: if 1) the angle between eye centers was greater than fifteen degrees, or 2) the interocular distance (number of pixels between eye centers) was less than one fifth of the image width, the located features were discarded.               Following the test of these conditions, a while loop conditioned on successful eye detection was used to refine the parameters of the detect function when eyes were not found. If this too proved unsuccessful, the image was stored for manual detection.
                
                When both eyes were successfully found, figure 2(D), we captured the coordinate location of the right $(x_r,y_r)$ and left $(x_l, y_l)$ eye centers by calculating the center location of the new bounding boxes. These eye centers were crucial for future steps.

			\subsection{Rotation}
				 Given successful eye-detection, the image is rotated based on the angle between the eye centers, as illustrated in Figure \ref{pipeline2}(E). Rotating the image began with the eye centers $(x_r,y_r)$ and $(x_l, y_l)$. We added a conditional to ensure the left eye stored in $(x_l, y_l)$ was actually the left eye of the subject. $\theta$ was then calculated by subtracting the left eye from the right eye, $(x_r-x_l, y_r-y_l)$, yielding the displacement between the eyes. We had to convert $\theta$ to the complex plane to allow use of the numpy angle function. $\theta$ was then a parameter used in the getRotationMatrix2D OpenCV function to create the necessary transformation matrix, M. 
				
				\begin{verbatim}
					cv2.getRotationMatrix2D(center, theta, scale)
					center = center of rotation source image
					scale = scale factor
				\end{verbatim}
				
				\begin{center}
					\begin{equation*}
						\begin{bmatrix}
							\alpha & \beta & (1-\alpha)\cdot center.x - \beta \cdot center.y \\
							-\beta & \alpha & \beta \cdot center.x + (1-\alpha) \cdot center.y \\
						\end{bmatrix},
					\end{equation*}
					\noindent where, $\alpha = scale \cdot \cos(angle), \beta = scale \cdot \sin(angle)$.
				\end{center}
				
				After the transformation matrix is calculated, it is applied to each pixel in the source image to produce the necessary rotated image. The OpenCV function warpAffine reconstructs the source image by producing the desired rotated image, {\it dst}:
                
				\begin{verbatim}
					cv2.warpAffine(src, M, (column, row))
				\end{verbatim}
				\begin{equation*}
					dst(x,y)=src(M_{11}x+M_{12}y+M_{13}, M_{21}x+M_{21}y+M_{21}).
				\end{equation*}

\subsection{Face and Eye Re-detection}
Following rotation, the face and eyes are re-detected as above, as illustrated in Figures \ref{pipeline1} and \ref{pipeline2}(E). Images with unsuccessfully detected faces were stored in a "fnf\_r" (face not found, rotated image) folder for manual detection later on. Images with undetected eyes were stored similarly, in a "enf\_r" (eyes not found, rotated image) folder.

\subsection{Cropping and Scaling}
After the eye centers were successfully re-located in the rotated image, a new bounding box for the face was determined based on the interocular distance. This step ultimately found the white bounding box in Figure \ref{pipeline2}(F) for the cropped image based on the interocular distance. Eye centers were relocated in the rotated image using the Haar cascade in section 2.3. The new eye center coordinates gave the interocular distance used for defining the height and width of the new bounding box. Finding the upper left corner of the box began with subtracting one interocular distance from the x-value midpoint of the eyes. The y-coordinate was calculated by adding four fifths times the interocular distance to the current eye height $(x=eyemidpoint-interoc, y=eyeheight+0.8\cdot interoc)$. [Note: 0.8 was chosen as the best option for capturing an appropriate region of the face]. The final bounding box was a slice of the rotated image with proportions 2:2.35 times the interocular distance.
				
The image was then cropped according to this frame and scaled down to 70 pixels tall by 60 pixels wide, 
using the following OpenCV function:
				\begin{verbatim}
					cv2.resize(cropped_img, (60, 70))
				\end{verbatim}
			
If an image could not be reduced to this size (i.e. something went wrong earlier on), it was stored in another folder for manual detection.                      
            
\subsection{Manual Pre-processing}
The images in which the face or eyes were not successfully detected were handled by manually clicking the eye centers of each subject. These new eye centers were then used for rotating the image and the rest of the pipeline followed suit. The images below in Figure \ref{errant detection} are examples of problem images that included errant detection.
\begin{figure}[ht]
\centering
    \begin{minipage}[t]{.3\linewidth}
      \centering
      \includegraphics[width=.5\textwidth]{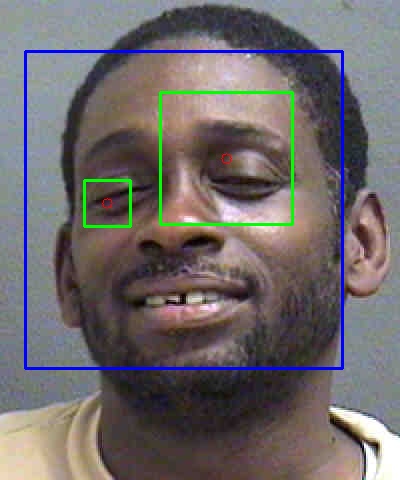}
    \end{minipage}%
    \hspace{-1cm}
    \begin{minipage}[t]{.3\linewidth}
      \centering
      \includegraphics[width=.5\textwidth]{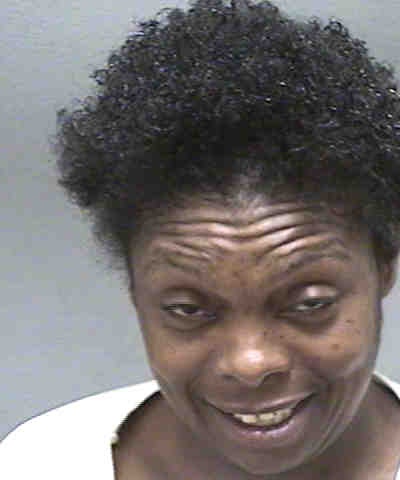}
    \end{minipage}%
    \hspace{-1cm}
    \begin{minipage}[t]{.3\linewidth}
      \centering
      \includegraphics[width=.5\textwidth]{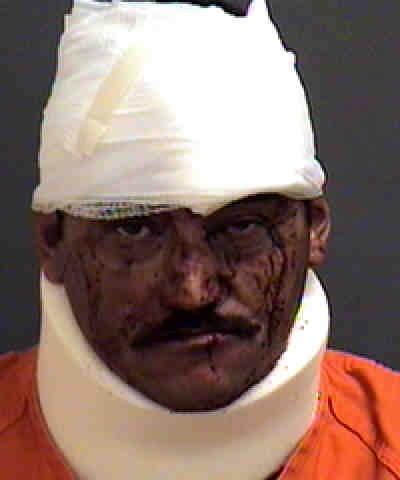}
    \end{minipage}
    \caption{Examples of problem images and errant detection}
    \label{errant detection}
  \end{figure}


\subsection{In Retrospect}

When performing the manual eye-detection, the images were rotated based on the eye centers. However, we avoided re-clicking the eyes in the rotated image by simply applying the rotation matrix to the coordinates of the eye centers in the unrotated image. Had this been recognized in the original code, the face and eye re-detection step could have been skipped entirely
. This would likely have drastically reduced run time and decreased the number of images necessitating manual eye detection.

		\section{Conclusion}
			When performing the manual eye-detection, the images were rotated based on the eye centers (as above). However, we avoided re-clicking the eyes in the rotated image by simply applying the rotation matrix to the coordinates of the eye centers in the unrotated image. Had this been recognized in the original code, the face and eye re-detection step could have been skipped entirely (effectively merging stages \textbf{2.3} and \textbf{2.4}). This would likely have drastically reduced run time and decreased the number of images necessitating manual eye detection.
		
\section{Acknowledgments}
This material is based in part upon work supported by the National Science Foundation under Grant Numbers DMS-1659288. Any opinions, findings, conclusions, or recommendations expressed in this material are those of the author(s) and do not necessarily reflect the views of the National Science Foundation.

\bibliographystyle{apa}
\bibliography{references}

\end{document}